# On-line Learning of Binary Lexical Relations Using Two-dimensional Weighted Majority Algorithms

Naoki Abe    Hang Li    Atsuyoshi Nakamura
Theory NEC Laboratory, RWCP*
c/o C & C Research Laboratories, NEC
4-1-1 Miyazaki Miyamae-ku, Kawasaki, 216 Japan.
{abe,lihang,atsu}@sbl.cl.nec.co.jp

## ABSTRACT

We consider the problem of learning a certain type of lexical semantic knowledge that can be expressed as a binary relation between words, such as the so-called sub-categorization of verbs (a verb-noun relation) and the compound noun phrase relation (a noun-noun relation). Specifically, we view this problem as an on-line learning problem in the sense of Littlestone's learning model [Lit88] in which the learner's goal is to minimize the total number of prediction mistakes. In the computational learning theory literature, Goldman, Rivest and Schapire [GRS93] and subsequently Goldman and Warmuth [GW93] have considered the on-line learning problem for binary relations $R : X \times Y \to \{0,1\}$ in which one of the domain sets $X$ can be partitioned into a relatively small number of types, namely clusters consisting of behaviorally indistinguishable members of $X$. In this paper, we extend this model and suppose that both of the sets $X, Y$ can be partitioned into a small number of types, and propose a host of prediction algorithms which are two-dimensional extensions of Goldman and Warmuth's weighted majority type algorithm proposed for the original model. We apply these algorithms to the learning problem for the 'compound noun phrase' relation, in which a noun is related to another just in case they can form a noun phrase together. Our experimental results show that all of our algorithms out-perform Goldman and Warmuth's algorithm. We also theoretically analyze the performance of one of our algorithms, in the form of an upper bound on the worst case number of prediction mistakes it makes.

## 1 Introduction

A major obstacle that needs to be overcome for the realization of a high quality natural language processing system is the problem of ambiguity resolution. It is generally acknowledged that some form of semantic knowledge is necessary for a successful solution to this problem. In particular, the so-called sub-categorization of verbs is considered essential, which asks *which verbs* can take *which nouns* as a subject, a direct object, or as any other grammatical role. A related form of knowledge is that of *which nouns* are likely to form compound noun phrases with *which other nouns*. These simple types of semantic knowledge can be expressed as a binary relation, or more in general an $n$-ary relation, between words. Since inputing such knowledge by hand is prohibitively expensive, automatic acquisition of such knowledge from large corpus data has become a topic of active research in natural language processing. (c.f.[PTL92, Per94])

In the computational learning theory literature, the problem of learning binary relations has been considered by Goldman et al [GRS93, GW93], in the on-line learning model of Littlestone [Lit88] and various extensions thereof. Note that a binary relation $R$ between sets $X$ and $Y$ can be thought of as a concept over the Cartesian product $X \times Y$, or a function from $X \times Y$ to $\{0,1\}$ defined by $R(x,y) = 1$ if and only if $R$ holds between $x \in X$ and $y \in Y$. Thus, Littlestone's on-line learning model for concepts can be directly adopted. Such a function can also be thought of as a matrix having value $R(x,y)$ at row $x$ and column $y$. Goldman et al assumed that the rows can be partitioned into a relatively small number of 'types', where any two rows $x_1, x_2 \in X$ are said to be of the same type if they are behaviorally indis-tinguishable, i.e. $R(x_1, y) = R(x_2, y)$ for all $y \in Y$. This is a natural assumption in our current problem setting, as indeed similar nouns such as 'man' and 'woman' seem to be indistinguishable with regard, for example, to the subject-verb relation. Under this assumption, the learning problem can be basically identified with the problem of discovering the proper clustering of nouns in an on-line fashion. Indeed the weighted majority type algorithm proposed by Goldman and Warmuth for this problem fits this intuition. (This is the algorithm 'Learn-Relation(0)' in [GW93], but in this paper we refer to it as WMP0.) Their algorithm keeps a 'weight' $w(x_1, x_2)$ representing the believed degree of similarity for any pair $x_1, x_2 \in X$, and at each trial predicts the label $R(x, y)$ by weighted majority vote among all $x' \in X$ such that it has already seen the correct label $R(x', y)$, each weighted according to $w(x, x')$. The weights are multiplicatively updated each time a mistake is made, reflecting whether $x'$ contributed positively or negatively to the correct prediction.

The above algorithm takes advantage of the similarities that exist within $X$, but does not make use of similarities that may exist within $Y$. In our current scenario, this may incurr a significant loss. In the subject-verb relation, not only the nouns but the verbs can also be classified into types. For example, the verbs 'eat' and 'drink' are sufficiently similar that they basically allow the same set of nouns as their subject. Motivated by this observation, in this paper we propose extensions of WMP0, called *2-dimensional* weighted majority prediction algorithms, which take advantage of the similarities that exist in both $X$ and $Y$.

We propose two basic variants of 2-dimensional weighted majority prediction algorithms, WMP1 and WMP2. Both of these algorithms make use of a weight $u(x_1, x_2)$ for each pair $x_1, x_2 \in X$ (called the 'row weights') and a weight $v(y_1, y_2)$ for each pair $y_1, y_2 \in Y$ (called the 'column weights'). WMP1 makes the prediction on input $(x, y) \in X \times Y$ by weighted majority vote over *all past examples*, with each pair weighted by the *product* of the corresponding row weight and column weight. It can thus make a rational prediction on a new pair $(i, j)$, even if both $i$ and $j$ are unseen in the past. The row weights are updated trusting the column weights and vice-versa. That is, after a prediction mistake occurs on $(i, j)$, each row weight $u(i, i')$ is multiplied by the ratio[1] between the sum of all column weights $v(j, j')$ for the columns $j'$ such that $M(i', j')$ contributed to the correct prediction for $(i, j)$, and the sum of $v(j, j')$ for the columns contributing to the wrong prediction. The more conservative of our two variants, WMP2, makes its predictions by majority vote over *only* the past examples in either *the same row* or in *the same column* as the current pair to be predicted. The weights are updated in a way similar to the update rule used in WMP0. We also use the following combination of these two algorithms, called WMP3. WMP3 predicts using the prediction method of WMP1, but updates its weights using the more conservative update rule of WMP2.

We apply all of these algorithms to on-line learning of lexical semantic knowledge, in particular to the problem of learning the 'compound noun phrase' relation, namely the binary relation between nouns in which a noun is related to another just in case they can together form a compound noun phrase. We extracted two-word compound noun phrases from a large 'tagged' corpus, and used them as training data for learning the relation restricted on those nouns that appear sufficiently frequently in the corpus.[2] Our experimental results indicate that our algorithms outperform WMP0 using weights on $X \times X$, which we call WMP0(X), and WMP0 with weights on $Y \times Y$, called WMP0(Y), as well as the weighted majority algorithm (exactly in the sense of [LW89]) which we call WMP4, using WMP0(X) and WMP0(Y) as sub-routines. These results also show that based on just 100 to 200 examples (representing 5 to 10 percent of the entire domain) our algorithms achieve about 80 to 85 per cent prediction accuracy on an unknown input.

We also theoretically analyze the performance of one of our algorithms. In particular, we give an upper bound on the worst-case number of mistakes made by WMP2 on any sequence of trials, in Littlestone's on-line learning model. The bound we obtain is $\frac{1}{k+l}(kl(m+n) + (ln + km)\sqrt{2(m+n)\log\frac{kl(m+n)}{ln+km}})$, where $n = |X|, m = |Y|$, $k$ is the number of row types, and $l$ is the number of column types. We note that this bound looks roughly like the weighted average of the bound shown by Goldman and Warmuth for WMP0(X), $km + n\sqrt{3m \log k}$, and that for WMP0(Y), $ln + m\sqrt{3n \log l}$, and thus tends to fall *in between* them.

---

[1] More precisely, we 'clump' this ratio between two constants, such as 0.5 and 2.0.

[2] Note that in any corpus data there are only positive examples, whereas the algorithms we propose here require the use of both positive and negative examples. We describe in Section 4 how we generate both positive and negative examples from a given corpus.

Finally, we tested all of our learning algorithms on randomly generated data for an artificially constructed target relation. The results of this experimentation confirm the tendency of our earlier experiment that WMP1, WMP2 and WMP3 outperform all of WMP0(X), WMP0(Y), and its weighted majority WMP4, apparently contradicting the above mentioned theoretical findings. Our interpretation of these results is that although in terms of the *worst case* mistake bounds, it is difficult to establish that our algorithms outperform the 1-dimensional algorithms, but in practice they seem to do better.

## 2 On-line Learning Model for Binary Relations

As noted in Introduction, a *binary relation* $R$ between sets $X$ and $Y$ is a concept over $X \times Y$, or equivalently a function from $X \times Y$ to $\{0, 1\}$ defined by $R(x, y) = 1$ if and only if $R$ holds between $x$ and $y$. In general, a learning problem can be identified with a subclass of the class of all concepts over a given domain. In this paper, we consider the subclass of all binary relations defined over finite sets $X \times Y$, in which both $X$ and $Y$ are classified into a relatively small number of 'types.' Formally, we say that a binary relation $R$ over $X \times Y$ is a $(k, l)$-relation, if there are at most $k$ row types and $l$ column types, namely $R$ satisfies the following conditions.

- There exist a partition $\mathcal{P} = \{P_i \subseteq X : i = 1, ..., k\}$ of $X$ such that $\forall P_i, i = 1, ..., k \, \forall x_1, x_2 \in P_i \, \forall y \in Y \, [R(x_1, y) = R(x_2, y)]$.
- There exist a partition $\mathcal{Q} = \{Q_j \subseteq: j = 1, ..., l\}$ of $Y$ such that $\forall Q_j, j = 1, ..., l \, \forall y_1, y_2 \in Q_j \, \forall x \in X \, [R(x, y_1) = R(x, y_2)]$.

Next, we describe the on-line learning model for binary relations. A learning session in this model consists of a sequence of trials. At each trial the learner is asked to predict the label of a previously unseen pair $(x, y) \in X \times Y$ based on the past examples. The learner is then presented with the correct label $R(x, y)$ as reinforcement. A learner is therefore a function that maps any finite sequence of labeled examples *and* a pair from $X \times Y$, to a prediction value, 0 or 1. A learner's performance is measured in terms of the total number of prediction mistakes it makes in the *worst case* over all possible instance sequences exhausting the entire domain, i.e. $X \times Y$. When the total number of mistakes made by a learning algorithm, when learning a target relation belonging to a given class, is always bounded above by a certain function, of various parameters quantifying the complexity of the learning problem, such as $|X|, |Y|, k$ and $l$, then we say that that function is a mistake bound for that algorithm and that class.

## 3 Two-dimensional Weighted Majority Prediction Algorithms

In this section, we give the details of all variants of 2-dimensional WMP algorithms informally described in Introduction, as well as the original 1-dimensional WMP algorithm of [GW93]. In the algorithm descriptions to follow, we use the following notation. We let $R$ denote the target relation to be learned, and $R(i, j)$ its label for $(i, j)$. We let $M$ denote the 'observation matrix' obtained from the past trials. That is, $M(i, j) = 1$ (or $M(i, j) = 0$) just in case $R(i, j) = 1$ (or $R(i, j) = 0$) has been observed in the past, and $M(i, j) =?$ indicates that $(i, j)$ has not been seen so far. When we write $M(i, j) \neq R(i', j')$, we mean that $M(i, j) \neq ?$ and $M(i, j) \neq R(i', j')$. Finally, we use WMP0(X) to denote WMP0 using weights between pairs of members of $X$, and WMP0(Y) to denote WMP0 using weights between pairs of members of $Y$.

**Algorithm** *WMP0(X)* [GW93]
(1-dimensional weighted majority prediction)
Initialize all weights $w(i, i')$ to 1
**Do Until** No more pairs are left to predict
Get a new pair $(i, j)$ and predict $R(i, j)$ as follows:
    **If** $\sum_{M(i', j)=1} w(i, i') > \sum_{M(i', j)=0} w(i, i')$
        then predict $R(i, j) = 1$
        else predict $R(i, j) = 0$
Get the correct label $R(i, j)$
**If** a prediction mistake is made
    then for all $i'$ such that $M(i', j) = R(i, j)$
    $w(i, i') := (2 - \gamma) \cdot w(i, i')$
    and for all $i'$ such that $M(i', j) \neq R(i, j)$
    $w(i, i') := \gamma \cdot w(i, i')$
**End Do**

**Algorithm** *WMP1*
(weighted majority over all past examples)

**For all** $i, j$, $u(i,i) := u_{init}; v(j,j) := v_{init}$
Initialize all other weights to 1
**Do Until** No more pairs are left to predict
Get a new pair $(i,j)$ and predict $R(i,j)$ as follows:
    **If** $\sum_{M(i',j')=1} u(i,i') \cdot v(j,j')$
        $> \sum_{M(i',j')=0} u(i,i') \cdot v(j,j')$
        **then** predict $R(i,j) = 1$
        **else** predict $R(i,j) = 0$
Get the correct label $R(i,j)$
**If** a prediction mistake is made
    **then** for all $i', j'$ update weights as follows
$$u^* := \max\{u_{low}, \min\{u_{up}, \frac{\sum_{M(i',j')=R(i,j)} v(j,j')}{\sum_{M(i',j') \neq R(i,j)} v(j,j')}\}\}$$
$$u(i,i') := u(i,i') \cdot u^*$$
$$v^* := \max\{v_{low}, \min\{v_{up}, \frac{\sum_{M(i',j')=R(i,j)} u(i,i')}{\sum_{M(i',j') \neq R(i,j)} u(i,i')}\}\}$$
$$v(j,j') := v(j,j') \cdot v^*$$
**For all** $i$, $u(i,i) := \max\{u_{init}, u_{up} \cdot u(i,i)\}$
**For all** $j$, $v(j,j) := \max\{v_{init}, v_{up} \cdot v(j,j)\}$
**End Do**

**Algorithm** *WMP2*
(weighted majority over same row and column)
Initialize all weights to 1
**Do Until** No more pairs are left to predict
Get a new pair $(i,j)$ and predict $R(i,j)$ as follows
    **If** $\sum_{M(i',j)=1} u(i,i') + \sum_{M(i,j')=1} v(j,j')$
        $> \sum_{M(i',j)=0} u(i,i') + \sum_{M(i,j')=0} v(j,j')$
        **then** predict $R(i,j) = 1$
        **else** predict $R(i,j) = 0$
Get the correct label $R(i,j)$
**If** a prediction mistake is made
    **then** for all $i', j'$ update weights as follows
    **If** $M(i',j) = R(i,j)$ **then** $u(i,i') := (2-\gamma)u(i,i')$
    **else if** $M(i',j) \neq R(i,j)$ **then** $u(i,i') := \gamma \cdot u(i,i')$
    **If** $M(i,j') = R(i,j)$ **then** $v(j,j') := (2-\gamma)v(j,j')$
    **else if** $M(i,j') \neq R(i,j)$ **then** $v(j,j') := \gamma \cdot v(j,j')$
**End Do**

**Algorithm** *WMP3*
(mixed strategy between WMP1 and WMP2)
Initialize all weights to 1
**Do Until** No more pairs are left to predict
    Predict with the prediction rule of WMP1
    Update the weights by the update rule of WMP2
**End Do**

**Algorithm** *WMP4*
(weighted majority over WMP0(X) and WMP0(Y))
Initialize weights $w_1$ and $w_2$ to 1
**Do Until** No more pairs are left to predict
Get a new pair $(i,j)$ and predict $R(i,j)$ as follows:
    **If** $\frac{w_1 \cdot WMP0(X) + w_2 \cdot WMP0(Y)}{w_1 + w_2} > \frac{1}{2}$
        **then** predict $R(i,j) = 1$
        **else** predict $R(i,j) = 0$
Get the correct label $R(i,j)$ and update weights
as follows
    **If** $R(i,j) \neq WMP0(X)$
        **then** $w_1 := \beta w_1$ and update weights of
        WMP0(X) according to WMP0
    **If** $R(i,j) \neq WMP0(Y)$
        **then** $w_2 := \beta w_2$ and update weights of
        WMP0(Y) according to WMP0
**End Do**

In the above description of WMP1, $u_{up}, u_{low}, v_{up}$ and $v_{low}$ are any reals satisfying $u_{up} > 1$, $u_{low} < 1$, $v_{up} > 1$ and $v_{low} < 1$, but we set $u_{up} = v_{up} = 2$ and $u_{low} = v_{low} = \frac{1}{2}$ in our experiments. We set $u_{init} = v_{init} = 10$ in our experiments. In WMP0 and WMP2, we set $\gamma = \frac{2\beta}{1+\beta}$ for some $\beta \in [0,1)$, so that we have $\gamma/(2-\gamma) = \beta$. In our experiments,[3] we used $\beta = \frac{1}{4}$. Finally, in WMP4, $\beta$ can be any real number in the range $(0,1)$, but in our experiments we set $\beta = \frac{1}{2}$.

## 4 Experimental Results

### 4.1 Learning Lexical Semantic Knowledge

We performed experiments on the problem of learning the 'compound noun phrase' relations. As training data, we used two-word compound noun phrases extracted from a large tagged corpus. The problem here is that although our learning algorithms make use of positive and negative examples, only positive examples are directly available in any corpus data. To solve this problem, we make use of the notion of 'association ratio,' which has been proposed and used by Church and Hanks [CH89] in the context of 'corpus-based' natural language processing. The association ratio between $x$ and $y$ quantifies the likelihood of co-occurrence of $x$ and $y$, and is defined as follows. (All logarithms are to the base 2 in this

---

[3] When we use WMP0 or WMP2 to predict a target relation which is 'pure' in the sense [GW93] that it is exactly a $(k,l)$-binary relation for some small $k$ and $l$, we can let $\beta = 0$. In practice, however, it is likely that the target relation is *almost* a $(k,l)$-binary relation with a few exceptions. When learning such a relation, setting $\beta = 0$ is too risky and it is better to use a more conservative setting, such as $\beta = \frac{1}{4}$.

paper.)

$$\log \frac{P(x,y)}{P(x)P(y)} \quad (1)$$

We wrote $P(x)$, $P(y)$ for the respective occurrence probability for $x$ and $y$, and $P(x,y)$ for the co-occurrence probability of $x$ and $y$. In the actual experiments, we used pairs of nouns with association ratio greater than 0.5 as positive examples, and those with association ratio less than -4.5 as negative examples.

We now give a detailed description of our experiments. We extracted approximately 80,000 two-word noun phrases from the Penn Tree Bank tagged corpus consisting of 120,000 sentences. We then performed our learning experiments focusing on the 53 most frequently appearing nouns on the left and the 40 most frequently appearing nouns on the right. We show the entire lists of these nouns in Figures 1 and 2. We then obtained positive and negative examples for these 53 × 40 pairs of nouns listed above from the corpus using association ratio, as described earlier in this section. There were 512 of these. Figure 3 shows several of these examples chosen arbitrarily from the 512 examples, paired with their association ratios.

In our experiments, we evaluated various prediction algorithms by the number of prediction mistakes they make on the training data obtained in the manner just described. More specifically, using a random number generator, we obtained ten distinct random permutations of the 512 training data, and we tested and compared the number of prediction mistakes made by WMP1 through WMP4 as well as WMP0.

The results of this experiment are shown in Figure 4. Figure 4(a) shows how the *cumulative prediction accuracy*, i.e. the number of mistakes made up to that point divided by the number of trials, changes at various stages of a learning session, averaged over the ten sessions. Figure 4(b), on the other hand, plots (the approximation of) the *instantaneous prediction accuracy* achieved at various stages in a learning session, again averaged over the ten sessions. More precisely, the value plotted at each trial is the average percentage of correct predictions in the last 50 trials (leading up to the trial in question).

Inspecting these experimental results reveals a certain definite tendency. That is, with respect to both the cumulative prediction accuracy (or equivalently the total number of prediction mistakes made), and the 'instantaneous' prediction accuracy, all of the algorithms we propose outperform WMP0(X), WMP0(Y) and their weighted majority. It is worth noting that the instantaneous prediction accuracy achieved by our algorithms after 100 trials is already about 80 per cent and after 200 trials reaches about 85 per cent, and then levels off. This seems to indicate that after seeing only 5 to 10 per cent of the entire domain, they achieve the level of generalization that is close to the best possible for this particular problem, which we suspect is quite noisy.

Examining the final settings of the weights, it did not appear as if our learning algorithms were discovering very clear clusters. Moreover, the final weight settings of WMP1 and WMP2 were not particularly correlated, even though their predictive performances were roughly equal. In Figure 5, we exhibit the final settings of the column weights in WMP1 between the noun 'stock' and some of the other column nouns, sorted in the decreasing order. Perhaps it makes sense that the weight between 'stock' and 'maker' is set small, for example, but in general it is hard to say that a proper clustering has been discovered. Interestingly, however, its predictive performance is quite satisfactory.

We feel that these results are rather encouraging, considering (i) that the target relation is most likely not a pure $(k,l)$-relation for reasonably small $k$ and $l$, and (ii) that among the nouns that were used in this experiment, there are not so many 'related' ones, since we chose the 40 (or 53) most frequently occurring nouns in a given corpus.

### 4.2 Simulation Experiments with Artificially Generated Data

We performed controlled experiments in which we tested all of our algorithms on artificially generated data. We used as the target relation a 'pure relation' defined over a domain of a comparable size to our earlier experiment (40 × 50), having 4 row types and 5 column types. In other words, the parameter setting we chose are $n = 40, m = 50, k = 4$, and $l = 5$. Each row and column type was equally sized (at 10). We tested our algorithms, plus WMP0(X), WMP0(Y), and WMP4 on ten randomly generated *complete* trial sequences, namely sequences of length 40 × 50. As before, Figure 6(a) shows the cumulative prediction accuracy (at various stages of a learning session) averaged over ten learning sessions, and

| interest   | percentage | money  | government | ad         | state     | trade    |
|------------|------------|--------|------------|------------|-----------|----------|
| trading    | consumer   | rate   | executive  | work       | operating | vice     |
| sale       | mortgage   | oil    | product    | stock      | security  | share    |
| insurance  | service    | bank   | auto       | law        | production| computer |
| company    | bond       | asset  | capital    | investment | industry  | future   |
| market     | program    | equity | exchange   | business   | food      | brokerage|
| fund       | junk       | drug   | tax        | debt       | revenue   | defense  |
| car        | price      | growth | credit     |            |           |          |

Figure 1: Nouns on the left hand side

| president | rate     | point    | fund    | official  | group    | yesterday |
|-----------|----------|----------|---------|-----------|----------|-----------|
| increase  | officer  | force    | law     | profit    | growth   | payment   |
| concern   | company  | market   | firm    | price     | manager  | agency    |
| issue     | stock    | index    | maker   | system    | line     | sale      |
| executive | contract | industry | analyst | operation | business | trader    |
| bond      | security | value    | gain    | share     |          |           |

Figure 2: Nouns on the right hand side

Figure 6(b) plots the average approximate instantaneous prediction accuracy, calculated using 50 most recent trials at each trial.

These results seem to indicate that, at least for pure relations with reasonable number of types, all our algorithms, WMP1, WMP2 and WMP3, outperform WMP0(X), WMP0(Y) and their weighted majority, confirming the tendency observed in our earlier experiment on lexical semantic knowledge acquisition. Moreover, the learning curves obtained for the simulation experiments are quite close to those for the earlier experiment.

Our algorithms achieve about 93 per cent cumulative prediction accuracy at the end of a learning session. This means that roughly $2000 \times 0.07 = 140$ mistakes were made in total. How does this compare with theoretical bounds on the number of mistakes for these algorithms ? In a companion paper [NA95], it is shown that a worst case number of mistakes for *any* algorithm learning a $(k,l)$-binary relation is at least $kl + (n - k)\log k + (m - l)\log l$. Plugging in the values $n = 40, m = 50, k = 4$, and $l = 5$, we obtain 216.1. So our algorithms seem to perform in practice even better than the theoretically best possible *worst case* behavior by any algorithm. In the next section, we show for WMP2 the mistake bound

$$\frac{1}{k+l}\left(kl(m+n) + (ln + km)\sqrt{2(m+n)\log\frac{kl(m+n)}{ln+km}}\right)$$

which upon substitution of the concrete values becomes 907.76. The bounds due to Goldman and Warmuth [GW93] on WMP0(X) and WMP0(Y), $km + n\sqrt{3m\log k}$ and $ln + m\sqrt{3n\log l}$, come out to be 892.8 and 1034.6, respectively. Although these bounds seem to be all gross over-estimates of the number of mistakes in the typical situation we have here,[4] the tendency is clear. The bound for WMP2 is worse than the better of the bounds for WMP0(X) and WMP0(Y). In our experiments, this is not the case and our 2-dimensional extensions out-perform both WMP0(X) and WMP0(Y). Our feeling is that this does not necessarily mean that our mistake bound can be improved drastically. Rather, these findings seem to cry for the need of theoretical analysis of typical behavior of these algorithms, perhaps in some form of average case analysis.

## 5 Theoretical Performance Analysis

In this section, we prove the following mistake bound for WMP2. As we noted in Introduction, our upper bound looks roughly like the weighted average of the bounds of [GW93] for WMP0(X) and WMP0(Y), and thus tends to be *in between* the two bounds. We add that we have not been able to prove a rigorous mistake bound for WMP1. We expect that in fact no non-trivial *worst case* mistake bound for WMP1 exists.

---

[4] It should be noted that these bounds become much more sensible for larger values of $n, m, k, l$.

| | | |
|---|---|---|
| market manager -4.769645 | ad industry 1.838923 | equity issue 1.446904 |
| production growth 1.928957 | future contract 4.337400 | vice concern -4.715181 |
| capital gain 5.032259 | government security 1.997249 | company official 4.345481 |
| service price -4.651155 | food issue 1.028839 | sale growth 5.309577 |
| mortgage payment 2.601217 | industry market -5.716859 | equity group 0.860507 |
| auto maker 3.871484 | law price -4.998001 | service firm 0.562419 |
| future company -4.999019 | program increase 0.976705 | revenue bond 3.955944 |
| insurance president -5.614235 | product group 0.926659 | exchange president -5.099226 |

Figure 3: Part of the training data

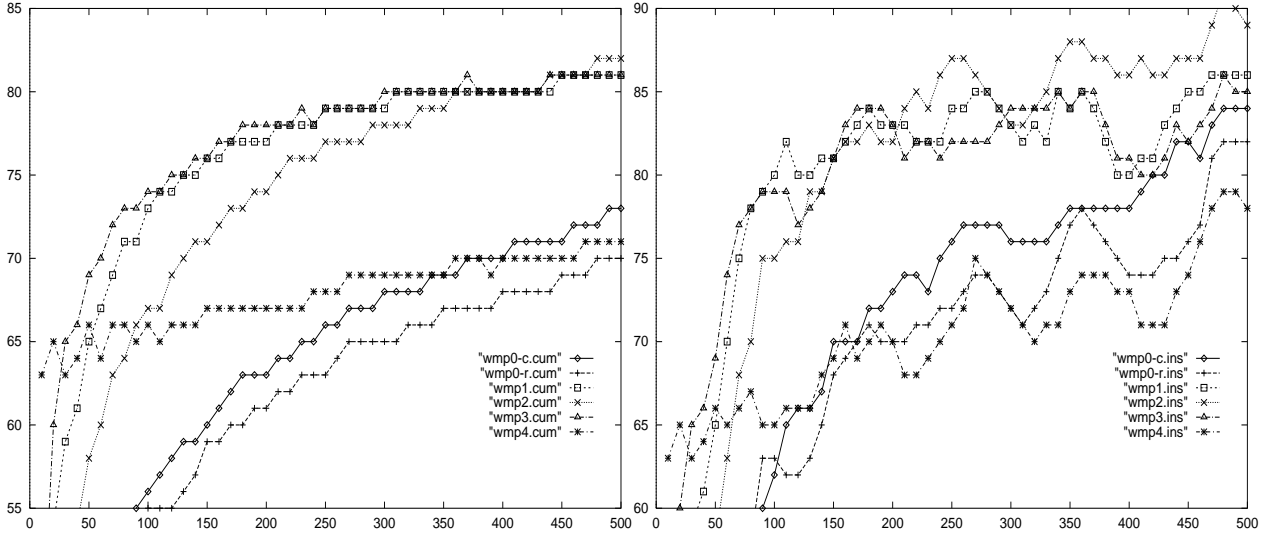

Figure 4: (a) Average cumulative prediction accuracy and (b) Average instantaneous prediction accuracy

**Theorem 5.1** *When learning a (k,l)-binary relation, Algorithm WMP2 makes at most*

$$\frac{1}{k+l}\left(kl(m+n)+(ln+km)\sqrt{2(m+n)\log\frac{kl(m+n)}{ln+km}}\right)$$

*mistakes in the worst case, provided* $k, l \geq 2$.

(Proof) We need the following definitions and notation. Let $n_p$ denote the number of rows of type $p$ and let $m_q$ denote the number of columns of type $q$. Let $\mu_p^r$ denote the number of mistakes made in row type $p$, and let $\mu_q^c$ denote the number of mistakes made in column type $q$. We then let $\mu$ denote the total number of mistakes, i.e., $\mu = \sum_{p=1}^{k}\mu_p^r = \sum_{q=1}^{l}\mu_q^c$. We write $\mathcal{E}_p^r$ for the set of all edges between two rows of type $p \in \{1,...,k\}$, and $\mathcal{E}_q^c$ for the set of all edges between two columns of type $q \in \{1,...,l\}$. We write $e_{i_1 i_2}^r$ for the edge between row $i_1$ and row $i_2$, and $e_{j_1 j_2}^c$ the edge between column $j_1$ and column $j_2$. Extending the notion of 'force' used in the proof of Theorem 4 in [GW93], for each prediction mistake made, say in predicting $(i,j)$, we define the *row force* of the mistake to be the number of rows $i'$ of the same type as $i$ for which $R(i',j)$ was known at the time of the prediction. Let $F_p^r$ denote the sum of the row forces of all mistakes made in row type $p$. We define the *column force* of a mistake analogously, and let $F_q^c$ denote the sum of column forces of mistakes made in column type $q$.

The theorem is proved using the following two lemmas.

| stock-index 2.0 | stock-executive 1.0 | stock-share 1.0 | stock-security 0.74 | stock-trader 0.59 |
| stock-sale 0.55 | stock-system 0.52 | stock-contract 0.5 | stock-industry 0.5 | stock-analyst 0.5 |
| stock-operation 0.5 | stock-business 0.5 | stock-bond 0.5 | stock-value 0.5 | stock-gain 0.5 |
| stock-line 0.25 | stock-maker 0.22 | | | |

Figure 5: WMP1's weights between 'stock' and other column nouns

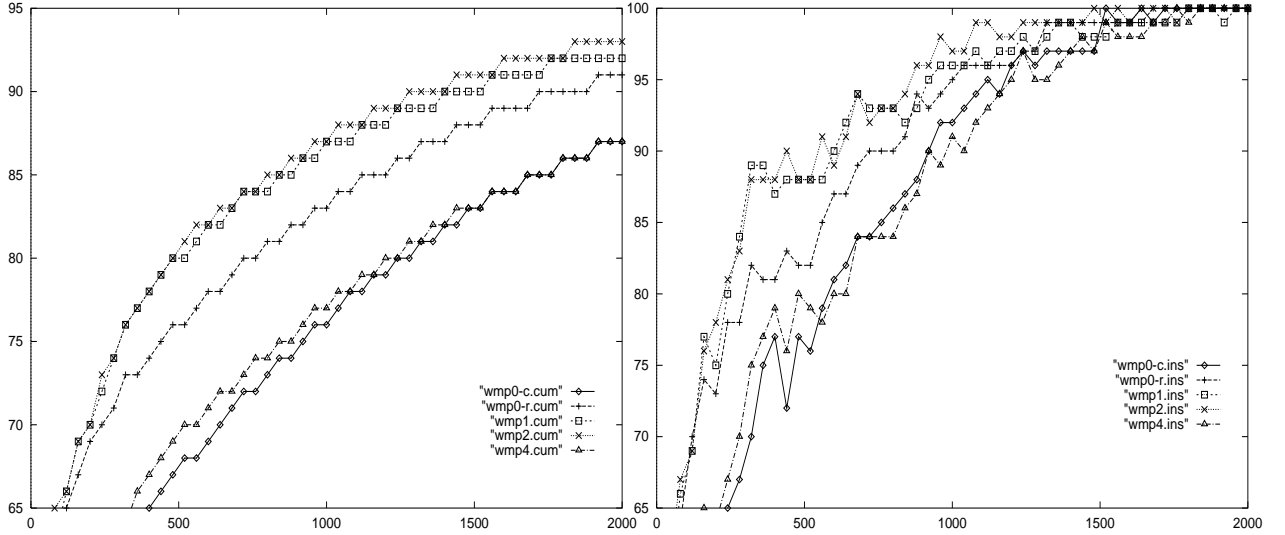

Figure 6: (a) Average cumulative prediction accuracy and (b) Average instantaneous prediction accuracy

**Lemma 5.1** *For each* $1 \leq p \leq k, 1 \leq q \leq l$,
$$F_p^r + F_q^c \leq (|\mathcal{E}_p^r| + |\mathcal{E}_q^c|) \log \frac{n(n-1) + m(m-1)}{2(|\mathcal{E}_p^r| + |\mathcal{E}_q^c|)}.$$

(Proof) The following two inequalities can be shown in a similar manner to the proofs of Lemma 1 and Lemma 2 in [GW93]:
$$F_p^r + F_q^c \leq \sum_{e \in \mathcal{E}_p^r \cup \mathcal{E}_q^c} \log w(e),$$
$$\sum_{e \in \mathcal{E}_p^r \cup \mathcal{E}_q^c} w(e) \leq \frac{n(n-1) + m(m-1)}{2}.$$

The lemma now follows easily from these two inequalities and Jensen's inequality. □

The following analogues for inequality (4) in [GW93] for the row and column forces can be readily shown.

**Lemma 5.2** *For each* $1 \leq p \leq k, 1 \leq q \leq l$, *both of the following hold.*
$$F_p^r \geq \frac{(\mu_p^r)^2}{2m} - \mu_p^r + \frac{m}{2} \ , \ F_q^c \geq \frac{(\mu_q^c)^2}{2n} - \mu_q^c + \frac{n}{2}$$

Now from Lemma 5.1 and Lemma 5.2 we obtain
$$\frac{(\mu_p^r - m)^2}{m} + \frac{(\mu_q^c - n)^2}{n}$$
$$\leq 2(|\mathcal{E}_p^r| + |\mathcal{E}_q^c|) \log \frac{n(n-1) + m(m-1)}{2(|\mathcal{E}_p^r| + |\mathcal{E}_q^c|)}.$$

Thus the following inequality follows.
$$(\mu_p^r - m) + (\mu_q^c - n)$$
$$\leq \sqrt{2(m+n)(|\mathcal{E}_p^r| + |\mathcal{E}_q^c|) \log \frac{n(n-1) + m(m-1)}{2(|\mathcal{E}_p^r| + |\mathcal{E}_q^c|)}}.$$

By summing the above over $k, l$ we get
$$(k+l)\mu \leq kl(m+n) + \sum_{p=1}^{k} \sum_{q=1}^{l}$$
$$\sqrt{2(m+n)(|\mathcal{E}_p^r| + |\mathcal{E}_q^c|) \log \frac{n(n-1) + m(m-1)}{2(|\mathcal{E}_p^r| + |\mathcal{E}_q^c|)}}.$$

Since $f(x) = x\sqrt{\log \frac{c}{x^2}}$ is concave for constant $c > 0$, the following can be shown to hold, where we let $a = \sum_{p=1}^{k} \sum_{q=1}^{l} \sqrt{2(|\mathcal{E}_p^r| + |\mathcal{E}_q^c|)}$.

$$(k+l)\mu \leq kl(m+n) + \sqrt{m+n} \cdot kl \cdot$$
$$\frac{a}{kl}\sqrt{\log \frac{n(n-1) + m(m-1)}{(\frac{a}{kl})^2}}$$
$$= kl(m+n) + \sqrt{m+n} \cdot$$

$$a\sqrt{\log \frac{k^2 l^2 (n(n-1) + m(m-1))}{a^2}}$$
$$\leq kl(m+n) + \sqrt{m+n} \cdot$$
$$a\sqrt{\log \frac{k^2 l^2 (n+m)^2}{a^2}}$$

Since $a = \sum_{p=1}^{k} \sum_{q=1}^{l} \sqrt{n_p(n_p - 1) + m_q(m_q - 1)}$ $\leq \sum_{p=1}^{k} \sum_{q=1}^{l} (n_p + m_q) = ln + km$ and $f(x) = x\sqrt{\log \frac{c}{x^2}}$ is monotonically increasing for $x$ in the range $\frac{c}{x^2} \geq e$, for $k, l \geq 2$ we have

$$(k+l)\mu \leq kl(m+n) + (ln+km)\sqrt{2(m+n)\log \frac{kl(n+m)}{ln+km}}.$$

The theorem follows immediately from this. □

## 6 Concluding Remarks

We have presented 2-dimensional extensions of the weighted majority prediction algorithm of [GW93] for binary relations, and applied them to the problem of learning the 'compound noun phrase' relation. A common approach to this problem in natural language processing makes use of some *a priori* knowledge about the noun clusters, usually in the form of a thesaurus. (c.f. [Res92].) Our algorithms make no use of such knowledge. Another common approach is the statistical clustering approach (c.f. [PTL92]), which views the clustering problem as the maximum likelihood estimation of a word co-occurrence distribution. Such an approach is based on a sound theory of statistics, but is often computational intractable as the clustering problem is NP-complete even in the 1-dimensional case. Our formulation of this problem as an on-line learning problem of deterministic binary relations gives rise to algorithms that are especially simple and efficient. Our algorithms seem to somehow bypass having to explicitly solve the clustering problem, and yet achieve reasonably high predictive performance. Note also that our upper bound on the worst case number of mistakes made by WMP2 relies on no probabilistic assumption on the input data. In the future, we would like to apply our algorithms on other related problems, such as that of learning verb sub-categorization relations.

## Acknowledgement

We thank Mr. K. Kinashi and Mr. T. Futagami of NIS for their programming efforts. We also thank Mr. K. Nakamura and Mr. T. Fujita of NEC C & C Research Laboratories for their encouragement. We acknowledge the A.C.L. for providing the Penn Tree Bank tagged corpus (ACL DCI CD-ROM-1).

## References

[CH89] K. Church and P. Hanks. Word association norms, mutual information, and lexicography. In *Proc. 27th Annu. Meeting of the A. C. L.*, 1989.

[GRS93] S. A. Goldman, R. L. Rivest, and R. E. Schapire. Learning binary relations and total orders. *SIAM J. Comput.*, 22(5):1006–1034, October 1993.

[GW93] S. Goldman and M. Warmuth. Learning binary relations using weighted majority voting. In *Proc. 6th Annu. Workshop on Comput. Learning Theory*, pages 453–462. ACM Press, New York, NY, 1993.

[Lit88] N. Littlestone. Learning quickly when irrelevant attributes abound: A new linear-threshold algorithm. *Machine Learning*, 2:285–318, 1988.

[LW89] N. Littlestone and M. K. Warmuth. The weighted majority algorithm. In *Proc. 30th Annu. IEEE Sympos. Found. Comput. Sci.*, pages 256–261. IEEE Computer Society Press, Los Alamitos, CA, 1989.

[NA95] A. Nakamura and N. Abe. On-line learning of binary and n-ary relations over multi-dimensional clusters. In *Proc. 8th Annu. Workshop on Comput. Learning Theory*, 1995.

[Per94] F. Pereira. Frequencies v.s. biases: Machine learning problems in natural language processing. In *Proc. 11th Int'l Conf. on Machine Learning*, page 380, 1994.

[PTL92] F. Pereira, N. Tishby, and L. Lee. Distributional clustering of english words. In *Proc. 30th Meeting of the A.C.L.*, pages 183–190, 1992.

[Res92] P. Resnik. Semantic classes and syntactic ambiguity. In *Prof. ARPA Workshop on Human Language Technology*, 1992.